%% file: coref.tex
\documentclass[11pt,a4paper]{article}
\usepackage[hyperref]{acl2020}
\usepackage{times}
\usepackage{latexsym}

\aclfinalcopy 
\def\aclpaperid{1590} 

\newcommand{\tablesize}{\footnotesize}

\usepackage{latexsym}
\usepackage{xstring}
\usepackage{xspace}
\usepackage{graphicx}
\usepackage{mathtools}
\usepackage{tikz}
\usepackage{colortbl}
\usepackage{cancel}

\usepackage{styles/haldefs}
\usepackage{url}            
\usepackage{booktabs}       
\usepackage{amsfonts}       
\usepackage{nicefrac}       
\usepackage{microtype}      
\usepackage{calc}

\usepackage{subfig}
\usepackage{graphicx}

\usepackage{microtype}

\makeatletter
\DeclareRobustCommand\citepos{\begingroup\let\NAT@nmfmt\NAT@posfmt\NAT@swafalse\let\NAT@ctype\z@\NAT@partrue\@ifstar{\NAT@fulltrue\NAT@citetp}{\NAT@fullfalse\NAT@citetp}}
\let\NAT@orig@nmfmt\NAT@nmfmt
\def\NAT@posfmt#1{\NAT@orig@nmfmt{#1's}}
\makeatother

\definecolor{cellbgcolor}{gray}{0.9}
\newcolumntype{g}{>{\columncolor{cellbgcolor}}c}

\usepackage{mathptmx}

\newcommand{\corpusname}[1]{%
  \IfEqCase{#1}{%
    {LDC2001T02}{MUC7}%
    {LDC2003T13}{Zh-PB3}
    {LDC2005T09}{ACE04}%
    {LDC2005T33}{BBN}%
    {LDC2006T06}{ACE05}%
    {LDC2009T10}{LUAC}
    {LDC2009T26}{NXT}
    {LDC2013T12}{MASC3}
    {LDC2013T19}{Onto5}%
    {LDC2014T18}{ACE07}%
    {LDC2017T08}{PDC}%
    {LDC2017T10}{AMR2}%
    {GAP}{GAP}%
    {QB}{QB}%
  }[\textbf{\textcolor{red!50!black}{UNKNOWN(#1)}}]%
  }

\newcommand{\corpusfullname}[1]{%
  \IfEqCase{#1}{%
    {LDC2001T02}{Message Understanding Conference 7}%
    {LDC2003T13}{Chinese Propbank 3.0}%
    {LDC2005T09}{Automatic Content Extraction 2004}%
    {LDC2005T33}{BBN Pronoun Coreference Corpus}%
    {LDC2006T06}{Automatic Content Extraction 2005}%
    {LDC2009T10}{Language Understanding Annotation Corpus}%
    {LDC2009T26}{NXT Switchboard Annotations}%
    {LDC2013T12}{Manually Annotated Sub-Corpus}%
    {LDC2013T19}{Ontonotes Version 5}%
    {LDC2014T18}{Automatic Content Extraction 2007}%
    {LDC2017T08}{Phrase Detectives Corpus}%
    {LDC2017T10}{Abstract Meaning Representation v 2}%
    {GAP}{Gendered Ambiguous Pronouns}%
    {QB}{QuizBowl Coreferences}%
  }[\textbf{\textcolor{red!50!black}{UNKNOWN(#1)}}]%
  }

\usepackage{longtable}

\newcommand{\feminine}{\textrm{\textsc{fem}}\xspace}
\newcommand{\masculine}{\textrm{\textsc{masc}}\xspace}

\newcommand{\pron}[1]{\textsc{#1}\xspace}
\newcommand{\setting}[1]{\textsc{#1}\xspace}

\newcommand{\newdataset}{\textsc{GICoref}\xspace}
\newcommand{\MAP}{\textsc{MAP}\xspace}

\newcommand{\hlemph}[1]{\colorbox{yellow!60!white}{\emph{\!#1\!}}}
\newcommand{\lingex}[2]{\vspace{-0.3em}\enumsentence{\textsf{\tablesize #2} \label{ling:#1}} \vspace{-0.3em}}
\newcommand{\lingref}[1]{\textsf{\tablesize(\ref{ling:#1})}}
\makeatletter
\def\enumsentence{\@ifnextchar[{\@enumsentence}{\refstepcounter{enums}\@enumsentence[\textsf{\footnotesize(\theenums)}]}}
\makeatother

\newcommand\cites[1]{\citeauthor{#1}'s\ (\citeyear{#1})}

\commentlevelDefault  

\title{Toward Gender-Inclusive Coreference Resolution} 

\author{Yang Trista Cao\\
	University of Maryland\\
  	\texttt{ycao95@cs.umd.edu}\\ \And
	Hal Daum\'e III\\
  	University of Maryland\\
  	Microsoft Research \\
  \texttt{me@hal3.name}}

\begin{document}

\maketitle
\begin{abstract}
  \input{abstract}
\end{abstract}
\fi

\hypersetup{colorlinks=true, linkcolor=darkpurple, citecolor=darkblue, filecolor=darkblue}


\section{Introduction} \label{sec:intro} \input{intro}

\section{Related Work} \label{sec:related} \input{related}\label{sec:relatedwork}

\section{Linguistic \& Social Gender} \label{sec:background}

\input{background}
\section{Bias in Human Annotation} \label{sec:datay} \input{datay}

\section{Bias in Model Specifications} \label{sec:model} \input{model}

\section{Discussion and Moving Forward} \label{sec:discussion} \input{discussion}

\section*{Acknowledgments}

The authors are grateful to a number of people who have provided pointers, edits, suggestions, and annotation facilities to improve this work: Lauren Ackerman, Cassidy Henry, Os Keyes, Chandler May, Hanyu Wang, and Marion Zepf, all contributed to various aspects of this work, including suggestions for data sources for the GI Coref dataset. We also thank the CLIP lab at the University of Maryland for comments on previous drafts.

\bibliography{styles/bibfile-shortname,styles/bibfile,trista,anth}
\bibliographystyle{acl_natbib}

\clearpage
\appendix
\onecolumn
\include{examples}

\end{document}

%% file: abstract.tex
Correctly resolving textual mentions of people fundamentally entails making inferences about those people.
Such inferences raise the risk of systemic biases in coreference resolution systems, including biases that can harm binary and non-binary trans and cis stakeholders.
To better understand such biases, we foreground nuanced conceptualizations of gender from sociology and sociolinguistics, and develop two new datasets for interrogating bias in crowd annotations and in existing coreference resolution systems.
Through these studies, conducted on English text, we confirm that without acknowledging and building systems that recognize the complexity of gender, we build systems that lead to many potential harms.


%% file: intro.tex


Coreference resolution---the task of determining which textual references resolve to the same real-world entity---requires making inferences about those entities.
Especially when those entities are people, coreference resolution systems run the risk of making unlicensed inferences, possibly resulting in harms either to individuals or groups of people.
Embedded in coreference inferences are varied aspects of gender, both because gender can show up explicitly (e.g., pronouns in English, morphology in Arabic) and because societal expectations and stereotypes around gender roles may be explicitly or implicitly assumed by speakers or listeners.
This can lead to significant biases in coreference resolution systems: cases where systems ``systematically and unfairly discriminate against certain individuals or groups of individuals in favor of others'' \citep[p. 332]{Fri:Nis:96}.

Gender bias in coreference resolution can manifest in many ways; work by \citet{rudinger-etal-2018-gender}, \citet{zhao-etal-2018-gender}, and \citet{webster-etal-2018-mind} focused largely on the case of \emph{binary} gender discrimination in trained coreference systems, showing that current systems over-rely on social stereotypes when resolving \pron{he} and \pron{she} pronouns\footnote{Throughout, we avoid mapping pronouns to a ``gender'' label, preferring to use the pronoun directly, include (in English) \pron{she}, \pron{he}, the non-binary use of singular \pron{they}, and neopronouns (e.g., \pron{ze/hir}, \pron{xey/xem}), which have been in usage since at least the 1970s \citep{bustillos2011pronoun,webster-pronouns, bradley2019, hord2016, spivak1997}.} (see \autoref{sec:related}).
Contemporaneously, critical work in Human-Computer Interaction has complicated discussions around gender in other fields, such as computer vision~\citep{keyes18misgendering, hamidi2018gender}. 

Building on both lines of work, and inspired by \cites{keyes18misgendering} study of vision-based automatic gender recognition systems, we consider gender bias from a broader conceptual frame than the binary ``folk'' model.
We investigate ways in which folk notions of gender---namely that there are two genders, assigned at birth, immutable, and in perfect correspondence to gendered linguistic forms---lead to the development of technology that is exclusionary and harmful of binary and non-binary trans and cis people.\footnote{Following~\citet{glaad2007trans}, transgender individuals are those whose gender differs from the sex they were assigned at birth. This is in opposition to cisgender individuals, whose assigned sex at birth happens to correspond to their gender. Transgender individuals can either be binary (those whose gender falls in the ``male/female'' dichotomy) or non-binary (those for which the relationship is more complex).} 
Addressing such issues is critical not just to improve the quality of our systems, but more pointedly to minimize the harms caused by our systems by reinforcing existing unjust social hierarchies~\citep{lambert2019gendered}. 

There are several stakeholder groups who may easily face harms when coreference systems is used~\citep{blodgett2020harms}. 
Those harms includes several possible harms, both allocational and representation harms \citep{barocas2017problem}, including quality of service, erasure, and stereotyping harms. 
Following \citepos{bender2019stakeholders} taxonomy of stakeholders and \citepos{barocas2017problem} taxonomy of harms, there are several ways in which trans exclusionary coreference resolution systems can cause harm:
\begin{itemize}[label=$\diamond$, leftmargin=*]
  \item \emph{Indirect: subject of query.} If a person is the subject of a web query, pages about xem may be missed if ``multiple mentions of query'' is a ranking feature, and the system cannot resolve xyr pronouns $\Rightarrow$ quality of service, erasure.
  \item \emph{Direct: by choice.} If a grammar checker uses coreference, it may insist that an author writing hir third-person autobiography is repeatedly making errors when referring to hirself $\Rightarrow$ quality of service, stereotyping, denigration.
  \item \emph{Direct: not by choice.} If an information extraction system run on r\'esum\'es relies on cisnormative assumptions, job experiences by a candidate who has transitioned and changed his pronouns may be missed $\Rightarrow$ allocative, erasure.
  \item \emph{Many stakeholders.} If a machine translation system uses discourse context to generate pronouns, then errors can results in directly misgendering subjects of the document being translated $\Rightarrow$ quality of service, denigration, erasure.
\end{itemize}

\noindent
To address such harms as well as understand where and how they arise, we need to complicate (a) what ``gender'' means and (b) how harms can enter into natural language processing (NLP) systems.
Toward (a), we begin with a unifying analysis (\autoref{sec:background}) of how gender is socially constructed, and how social conditions in the world impose expectations around people's gender.
Of particular interest is how gender is reflected in language, and how that both matches and potentially mismatches the way people experience their  gender in the world. 
Then, in order to understand social biases around gender, we find it necessary to consider the different ways in which gender can be realized linguistically, breaking down what previously have been considered ``gendered words'' in NLP papers into finer-grained categories that have been identified in the sociolinguistics literature of lexical, referential, grammatical, and social gender.


Toward (b), we focus on how bias can enter into two stages of machine learning systems: data annotation (\autoref{sec:datay}) and model definition (\autoref{sec:model}). We construct two new datasets: (1) MAP (a similar dataset to GAP~\citep{webster-etal-2018-mind} but without binary gender constraints) on which we can perform counterfactual manipulations and (2) GICoref (a fully annotated coreference resolution dataset written by and about trans people).\footnote{Both datasets are released under a BSD license at \href{https://github.com/TristaCao/into_inclusivecoref}{\texttt{github.com/TristaCao/into\_inclusivecoref}} with corresponding datasheets~\citep{gebru2018datasheets}.}
In all cases, we focus largely on harms due to over- and under-representation~\citep{kay2015unequal}, replicating stereotypes~\citep{sweeney2013discrimination,caliskan2017semantics} (particular those that are cisnormative and/or heteronormative), and quality of service differentials~\cite{buolamwini2018gender}.


\noindent
\paragraph{The primary contributions of this paper are:}
(1) Connecting existing work on gender bias in NLP to sociological and sociolinguistic conceptions of gender to provide a scaffolding for future work on analyzing ``gender bias in NLP'' (\autoref{sec:background}).
(2) Developing an ablation technique for measuring gender bias in coreference resolution annotations, focusing on the \emph{human} bias that can enter into annotation tasks (\autoref{sec:datay}).
(3) Constructing a new dataset, the Gender Inclusive Coreference dataset (\newdataset), for testing performance of coreference resolution systems on texts that discuss non-binary and binary transgender people (\autoref{sec:model}).

%% file: related.tex


There are four recent papers that consider gender bias in coreference resolution systems. \citet{rudinger-etal-2018-gender} evaluates coreference systems for evidence of \emph{occupational stereotyping}, by constructing Winograd-esque \citep{levesque2012winograd} test examples.
They find that humans can reliably resolve these examples, but systems largely fail at them, typically in a gender-stereotypical way. In contemporaneous work, \citet{zhao-etal-2018-gender} proposed a very similar, also Winograd-esque scheme, also for measuring gender-based occupational stereotypes. In addition to reaching similar conclusions to \citet{rudinger-etal-2018-gender}, this work also used a similar ``counterfactual'' data process as we use in \autoref{sec:ablation} in order to provide additional training data to a coreference resolution system. \citet{webster-etal-2018-mind} produced the GAP dataset for evaluating coreference systems, by specifically seeking examples where ``gender'' (left underspecified) could \emph{not} be used to help coreference. 
They found that coreference systems struggle in these cases, also pointing to the fact that some success of current coreference systems is due to reliance on (binary) gender stereotypes. 
Finally, \citet{lauren2018} presents an alternative breakdown of gender than we use (\autoref{sec:background}), and proposes matching criteria for modeling coreference resolution linguistically, taking a trans-inclusive perspective on gender.

Gender bias in NLP has been considered more broadly than just in coreference resolution, including, 
natural language inference \citep{rudinger2017social}, 
word embeddings \citep[e.g.,][]{bolukbasi2016man,romanov2019name,gonen2019lipstick},
sentiment analysis \citep{kiritchenko-mohammad-2018-examining},
machine translation \citep{font2019equalizing,prates2019assessing,wals-101,
frank04gender,wandruszka69sprachen,nissen02aspects,doleschal2001doing}, among many others~\citep[inter alia]{blodgett2020harms}.
Gender is also an object of study in gender recognition systems \citep{hamidi2018gender}.
Much of this work has focused on gender bias with a (usually implicit) binary lens, an issue which was also called out recently by \citet{larson2017gender} and \citet{may2019}.

%% file: background.tex
The concept of gender is complex and contested, covering (at least) aspects of a person's internal experience, how they express this to the world, how social conditions in the world impose expectations on them (including expectations around their sexuality), and how they are perceived and accepted (or not).
When this complex concept is realized in language, the situation becomes even more complex: linguistic categories of gender do not even remotely map one-to-one to social categories.
As observed by \citet{bucholtz99gender}:

\renewenvironment{quote}
  {\list{}{\rightmargin=0.5cm \leftmargin=0.5cm}%
   \item\relax\vspace{-0.5em}}
  {\endlist\vspace{-0.5em}}

\begin{quote} 
\emph{``Attempts to read linguistic structure directly for information about social gender are often misguided.''}
\end{quote}

\noindent
For instance, when working in a language like English which formally marks gender on pronouns, it is all too easy to equate ``recognizing the pronoun that corefers with this name'' with ``recognizing the real-world gender of referent of that name.''

Furthermore, despite the impossibility of a perfect alignment with linguistic gender, it is generally clear that an incorrectly gendered reference to a person (whether through pronominalization or otherwise) can be highly problematic ~\citep{johnsonmisgendering, mclemoremisgendering}.
This process of \emph{misgendering} is problematic for both trans and cis individuals to the extent that transgender historian \citet{stryker08transgender} writes:

\begin{quote} 
  \emph{``[o]ne's gender identity could perhaps best be described as how one feels about being referred to by a particular pronoun.''}
\end{quote}





\subsection{Sociological Gender} \label{sec:worldgender}


Many modern trans-inclusive models of gender recognize that \textit{gender} encompasses many different aspects.
These aspects include the experience that one has of gender (or lack thereof),
the way that one expresses one's gender to the world,
and the way that normative social conditions impose gender norms, typically as a dichotomy between masculine and feminine roles or traits~\citep{kramarae85feminist,west87doing,butler90gender,risman09doing,serano2007whipping}.
Gender self-determination, on the other hand, holds that each person is the ``ultimate authority'' on their own gender identity~\citep{zimman2019, stanley2014}, with \citet{zimman2019} further arguing the importance of the role language plays in that determination.

Such trans-inclusive models deconflate anatomical and biological traits and the sex that a person had assigned to them at birth from one's gendered position in society;
this includes intersex people, whose anatomical/biological factors do not match the usual designational criteria for either sex.
Trans-inclusive views typically recognize that gender exists beyond the regressive ``female''/``male'' binary\footnote{Some authors use female/male for sex and woman/man for gender; we do not need this distinction (which is itself contestable) and use female/male for gender.}; additionally, one's gender may shift by time or context (often ``genderfluid''), and  some people do not experience gender at all (often ``agender'')~\citep{kessler78gender,schilt09doing,darwin17doing,richards17genderqueer}.
In \autoref{sec:model} we analyze the degree to which NLP papers make trans-inclusive or trans-exclusive assumptions. 

\textbf{Social gender} refers to the imposition of gender roles or traits based on normative social conditions~\citep{kramarae85feminist}, which often includes imposing a dichotomy between feminine and masculine (in behavior, dress, speech, occupation, societal roles, etc.).~\citet{lauren2018} highlights a highly overlapping concept, ``bio-social gender'', which consists of gender role, gender expression, and gender identity.
Taking gender role as an example, upon learning that a nurse is coming to their hospital room, a patient may form expectations that this person is likely to be ``female,'' and may generate expectations around how their face or body may look, how they are likely to be dressed, how and where hair may appear, how to refer to them, and so on. 
This process, often referred to as \emph{gendering}~\citep{serano2007whipping} occurs both in real world interactions, as well as in purely linguistic settings (e.g., reading a newspaper), in which readers may use social gender clues to assign gender(s) to the real world people being discussed.

\subsection{Linguistic Gender} \label{sec:linggender}

Our discussion of linguistic gender largely follows~\citep{corbett91gender,ochs92indexing,craig94classifier,wals-30,hellinger15gender, fuertes2007corpus}, departing from earlier characterizations that postulate a direct mapping from language to gender~\citep{lakoff75language,silverstein79language}.
Our taxonomy is related but not identical to \citep{lauren2018}, which we discuss in \autoref{sec:relatedwork}.

\textbf{Grammatical gender}, similarly defined in \citet{lauren2018}, is nothing more than a classification of nouns based on a principle of \emph{grammatical agreement}.
In ``gender languages'' there are typically two or three grammatical genders that have, for animate or personal references, considerable correspondence between a \feminine (resp. \masculine) grammatical gender and referents with female- (resp. male-)\footnote{One difficulty in this discussion is that linguistic gender and social gender use the terms ``feminine'' and ``masculine'' differently; to avoid confusion, when referring to the linguistic properties, we use \feminine and \masculine.} social gender.
In comparison, ``noun class languages'' have no such correspondence, and typically many more classes.
Some languages have no grammatical gender at all; English is generally seen as one~\citep{nissen02aspects, baron1971} (though this is contested~\citep{bjorkman2017}).

  
\textbf{Referential gender} (similar, but not identical to ~\cites{lauren2018}  ``conceptual gender") relates linguistic expressions to extra-linguistic reality, typically identifying referents as ``female,'' ``male,'' or ``gender-indefinite.''
Fundamentally, referential gender only exists when there is an entity being referred to, and their gender (or sex) is realized linguistically. The most obvious examples in English are gendered third person pronouns (\pron{she}, \pron{he}), including neopronouns (\pron{ze}, \pron{em}) and singular \pron{they}\footnote{People's mental acceptability of singular \pron{they} is still relatively low even with its increased usage~\citep{prasad_morris_2020}, and depends on context \citep{conrod2018singular}.}, but also includes cases like ``policeman'' when the intended referent of this noun has social gender ``male'' (though not when ``policeman'' is used non-referentially, as in ``every policeman needs to hold others accountable'').

\textbf{Lexical gender} refers to an extra-linguistic properties of female-ness or male-ness in a \emph{non-referential} way, as in terms like ``mother'' as well as gendered terms of address like ``Mrs.''
Importantly, lexical gender is a property of the linguistic unit, \emph{not} a property of its referent in the real world, which may or may not exist.
For instance, in ``Every son loves his parents'', there is no real world referent of ``son'' (and therefore no \emph{referential} gender), yet it still (likely) takes \textsc{his} as a pronoun anaphor because ``son'' has lexical gender \masculine.

\setlength{\fboxsep}{0.3pt}
\newcommand{\mystrut}{\rule[-.15\baselineskip]{0pt}{0.7\baselineskip}}
\newcommand{\txtmarka}[1]{\fcolorbox{purple!80!white}{purple!30!white}{\mystrut\textbf{#1}}$_{\textbf{a}}$}
\newcommand{\txtmarkb}[1]{\fcolorbox{orange!80!white}{orange!30!white}{\mystrut\textbf{#1}}$_{\textbf{b}}$}
\newcommand{\txtmarkc}[1]{\fcolorbox{green!80!white}{green!30!white}{\mystrut\textbf{#1}}$_{\textbf{c}}$}
\newcommand{\txtmarkd}[1]{\fcolorbox{blue!80!white}{blue!30!white}{\mystrut\textbf{#1}}$_{\textbf{d}}$}
\tikzstyle{nosep}=[inner sep=0pt, outer sep=0pt]
\newcommand{\hcancel}[1]{%
    \tikz[baseline=(tocancel.base)]{
        \node[nosep] (tocancel) {#1};
        \node[nosep, yshift=.2ex]  (from) at (tocancel.south west) {};
        \node[nosep, yshift=-.2ex] (to)   at (tocancel.north east) {};
        \draw[red] (from) -- (to);
    }%
}%

\newcommand{\txteditz}[3]{\mystrut{\hcancel{#1} $\xrightarrow{\text{(#3)}}$ \textbf{#2}}}

\newcommand{\txtedita}[2]{\fcolorbox{purple!80!white}{purple!30!white}{\txteditz{#1}{#2}{a}}}
\newcommand{\txteditb}[2]{\fcolorbox{orange!80!white}{orange!30!white}{\txteditz{#1}{#2}{b}}}
\newcommand{\txteditc}[2]{\fcolorbox{green!80!white}{green!30!white}{\txteditz{#1}{#2}{c}}}
\newcommand{\txteditd}[2]{\fcolorbox{blue!80!white}{blue!30!white}{\txteditz{#1}{#2}{d}}}

\begin{figure*}
  \tablesize\centering\footnotesize \sffamily

  \begin{tabular}{p{15.5cm}}
    \toprule
    \txteditd{Mrs.}{$\emptyset$}
   \txteditb{Rebekah Johnson Bobbitt}{M. Booth} was the younger 
   \txteditc{sister}{sibling} of 
   \txteditb{Lyndon B. Johnson}{T. Schneider}, 36th President of the United States. Born in 1910 in Stonewall, Texas,
   \txtedita{she}{they} worked in the cataloging department of the Library of Congress in the 1930s before 
   \txtedita{her}{their} 
   \txteditc{brother}{sibling} entered politics.
    \\
    \bottomrule
  \end{tabular}
  \caption{\label{tab:exampleablate} Example of applying \emph{all} ablation substitutions for an example context in the \MAP corpus. Each substitution type is marked over the arrow and separately color-coded.}
\end{figure*}

\subsection{Social and Linguistic Gender Interplays} \label{sec:interplays}




The relationship between these aspects of gender is complex, and none is one-to-one.
The referential gender of an individual (e.g., pronouns in English) may or may not match their social gender and this may change by context.
This can happen in the case of people whose everyday life experience of their gender fluctuates over time (at any interval), as well as in the case of drag performers (e.g., some men who perform drag are addressed as \pron{she} while performing, and \pron{he} when not \citep{drag2017}).
The other linguistic forms of gender (grammatical, lexical) also need not match each other, nor match referential gender~\citep{hellinger15gender}.

Social gender (societal expectations, in particular)  captures the observation that upon hearing ``My cousin is a librarian'', many speakers will infer ``female'' for ``cousin'', because of either an entailment of ``librarian'' or some sort of probabilistic inference~\citep{lyons77semantics}, but not based on either grammatical gender (which does not exist in English) or lexical gender.
We focus on English, which has no grammatical gender, but does have lexical gender.
English also marks referential gender on singular third person pronouns.

Below, we use this more nuanced notion of different types of gender to inspect how bias play out in coreference resolution systems. These biases may arise in the context of any of these notions of gender, and we encourage future work to extend care over and be explicit about  what notions of gender are being utilized and when.

%% file: datay.tex
A possible source of bias in coreference systems comes from human annotations on the data used to train them. Such biases can arise from a combination of (possibly) underspecified annotations guidelines and the positionality of annotators themselves.
In this section, we study how different aspects of linguistic notions impact an annotator's judgments of anaphora. This parallels \citet{lauren2018} linguistic analysis, in which a Broad Matching Criterion is proposed, which posits that ``matching gender requires at least one level of the mental representation of gender to be identical to the candidate antecedent in order to match." 

Our study can be seen as evaluating which conceptual properties of gender are most salient in human judgments.
We start with natural text in which we can cast the coreference task as a binary classification problem (``which of these two names does this pronoun refer to?'') inspired by~\citet{webster-etal-2018-mind}.
We then generate ``counterfactual augmentations'' of this dataset by ablating the various notions of linguistic gender described in \autoref{sec:linggender}, similar to \citet{zmigrod2019counterfactual}.
We finally evaluate the impact of these ablations on human annotation behavior to answer the question: which forms of linguistic knowledge are most essential for human annotators to make consistent judgments. 
See \autoref{ex:biasexs} for examples of how linguistic gender may be used to infer social gender.

\subsection{Ablation Methodology} \label{sec:ablation}

In order to determine \emph{which} cues annotators are using and the \emph{degree} to which they use them, we construct an ablation study in which we hide various aspects of gender and evaluate how this impacts annotators' judgments of anaphoricity.
We construct binary classification examples taken from Wikipedia pages, in which a single pronoun is selected, and two possible antecedent names are given, and the annotator must select which one.
We cannot use \citeauthor{webster-etal-2018-mind}'s GAP dataset directly, because their data is constrained that the ``gender'' of the two possible antecedents is ``the same''\footnote{It is unclear from the GAP dataset what notion of ``gender'' is used, nor how it was determined to be ``the same.''}; for us, we are specifically interested in how annotators make decisions even when additional gender information is available.
Thus, we construct a dataset called \emph{Maybe Ambiguous Pronoun} (\MAP) following \citeauthor{webster-etal-2018-mind}'s approach, but we do not restrict the two names to match gender. 

In ablating gender information, one challenge is that removing social gender cues (e.g., ``nurse'' tending female) is not possible because they can exist anywhere.
Likewise, it is not possible to remove syntactic cues in a non-circular manner.
For example in \lingref{syntax}, syntactic structure strongly suggests the antecedent of ``herself'' is ``Liang'', making it less likely that ``He'' corefers with Liang later (though it is possible, and such cases exist in natural data due either to genderfluidity or misgendering).
\lingex{syntax}{Liang saw herself in the mirror\dots \hlemph{He} \dots}

\noindent
Fortunately, it \emph{is} possible to enumerate a high coverage list of English terms that signal lexical gender: terms of address (Mrs., Mr.) and semantically gendered nouns (mother).\footnote{These are, however, sometimes complex. For instance, ``actress'' signals \emph{lexical} gender of female, while ``actor'' may signal \emph{social} gender of male and, in certain varieties of English, may also signal \emph{lexical} gender of male.} We assembled a list by taking many online lists (mostly targeted at English language learners), merging them, and manual filtering. The assembling process and the final list is published with the {\MAP} dataset and its datasheet.


To execute the ``hiding'' of various aspects of gender, we use the following substitutions:
\begin{enumerate}[label=(\alph*)]
\item $\neg$\setting{Pro}: Replace third person pronouns with gender neutral variants (\pron{they}, \pron{xey}, \pron{ze}).
\item $\neg$\setting{Name}: Replace names by random names with only a first initial and last name.
\item $\neg$\setting{Sem}: Replace semantically gendered nouns with gender-indefinite variants.
\item $\neg$\setting{Addr}: Remove terms of address.\footnote{An alternative suggested by Cassidy Henry that we did not explore would be to replace all with Mx. or Dr.}
\end{enumerate}
\noindent
See \autoref{tab:exampleablate} for an example of all substitutions.


We perform two sets of experiments, one following a ``forward selection'' type ablation (start with everything removed and add each back in one-at-a-time) and one following ``backward selection'' (remove each separately). Forward selection is necessary in order to de-conflate syntactic cues from stereotypes; while backward selection gives a sense of how much impact each type of gender cue has in the context of all the others.

We begin with \setting{Zero}, in which we apply all four substitutions.
Since this also removes gender cues from the pronouns themselves, an annotator cannot substantially rely on social gender to perform these resolutions.
We next consider adding back in the original pronouns (always \pron{he} or \pron{she} here), yielding $\neg$\setting{Name}$\neg$\setting{Sem}$\neg$\setting{Addr}.
Any difference in annotation behavior between \setting{Zero} and $\neg$\setting{Name}$\neg$\setting{Sem}$\neg$\setting{Addr} can only be due to social gender stereotypes.
The next setting, $\neg$\setting{Sem}$\neg$\setting{Addr} removes both forms of lexical gender (semantically gendered nouns and terms of address); differences between $\neg$\setting{Sem}$\neg$\setting{Addr} and $\neg$\setting{Name}$\neg$\setting{Sem}$\neg$\setting{Addr} show how much names are relied on for annotation.
Similarly, $\neg$\setting{Name}$\neg$\setting{Addr} removes names and terms of address, showing the impact of semantically gendered nouns,
and $\neg$\setting{Name}$\neg$\setting{Sem} removes names and semantically gendered nouns, showing the impact of terms of address.

In the backward selection case, we begin with \setting{Orig}, which is the unmodified original text.
To this, we can apply the pronoun filter to get $\neg$\setting{Pro}; differences in annotation between \setting{Orig} and $\neg$\setting{Pro} give a measure of how much \emph{any} sort of gender-based inference is used.
Similarly, we get $\neg$\setting{Name} by only removing names, which gives a measure of how much names are used (in the context of all other cues);
we get $\neg$\setting{Sem} by only removing semantically gendered words;
and $\neg$\setting{Addr} by only removing terms of address.


\begin{figure}[t]
  \centering
  \includegraphics[width = 7cm]{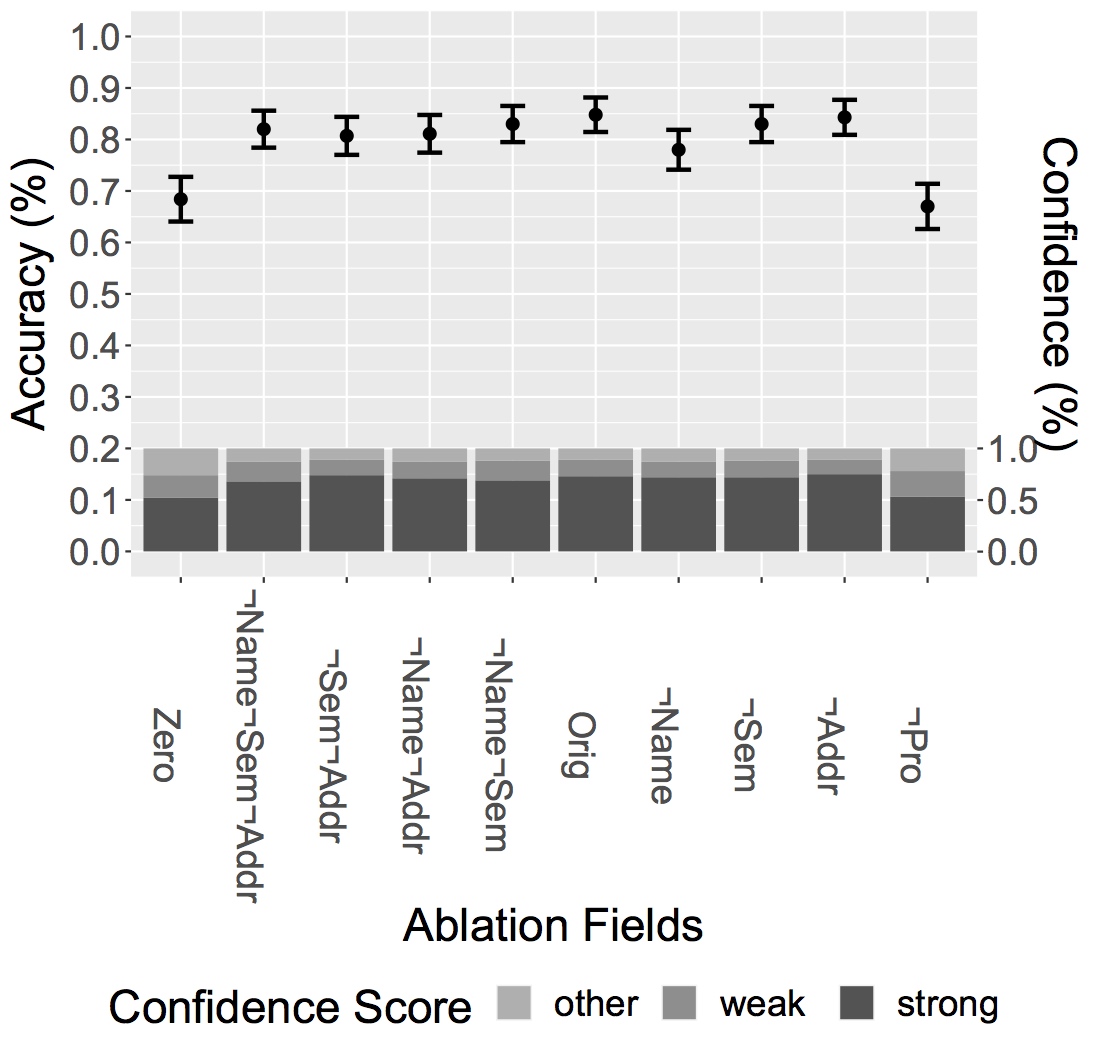}
  \caption{\label{fig:MAPhuman} Human annotation results for the ablation study on \MAP dataset. Each column is a different ablation, and the y-axis is the degree of \emph{accuracy} with 95\% significance intervals. Bottom bar plots are annotator certainties as how sure they are in their choices. }
\end{figure}

\subsection{Annotation Results} \label{sec:ablationresults}

We construct examples using the methodology defined above. We then conduct annotation experiments using crowdworkers on Amazon Mechanical Turk following the methodology by which the original GAP corpus was created\footnote{Our study was approved by the Microsoft Research Ethics Board. Workers were paid \$1 to annotate ten contexts (the average annotation time was seven minutes).}. Because we wanted to also capture uncertainty, we ask the crowdworkers how sure they are in their choices, between ``definitely'' sure, ``probably'' sure and ``unsure.''

\autoref{fig:MAPhuman} shows the human annotation results as binary classification accuracy for resolving the pronoun to the antecedent. We can see that removing pronouns leads to significant drop in accuracy. This indicates that gender-based inferences, especially social gender stereotypes, play the most significant role when annotators resolve coreferences. This confirms the findings of \citet{rudinger-etal-2018-gender} and \citet{zhao-etal-2018-gender} that human annotated data incorporates bias from stereotypes. 

Moreover, if we compare \setting{Orig} with columns left to it, we see that name is another significant cue for annotator judgments, while lexical gender cues do not have significant impacts on human annotation accuracies. This is likely in part due to the low appearance frequency of lexical gender cues in our dataset. Every example has pronouns and names, whereas 49\% of the examples have semantically gendered nouns but only 3\% of the examples include terms of address. We also note that if we compare $\neg$\setting{Name}$\neg$\setting{Sem}$\neg$\setting{Addr} to $\neg$\setting{Sem}$\neg$\setting{Addr} and $\neg$\setting{Name}$\neg$\setting{Addr}, accuracy drops when removing gender cues. Though the differences are not statistically significant, we did not expect the accuracy drop. 

Finally, we find annotators' certainty values follow the same trend as the accuracy: annotators have a reasonable sense of when they are unsure.
We also note that accuracy score are essentially the same for \setting{Zero} and \setting{$\neg$Pro}, which
suggests that once explicit binary gender is gone from pronouns, the impact of any other form of linguistic gender in annotator's decisions is also removed.

%% file: model.tex
In addition to biases that can arise from the data that a system is trained on, as studied in the previous section, bias can also come from how models are structured.
For instance, a system may fail to recognize anything other than a dictionary of fixed pronouns as possible referents to entities.
Here, we analyze prior work in models for coreference resolution in three ways.
First, we do a literature study to quantify how NLP papers discuss gender.
Second, similar to \citet{zhao-etal-2018-gender} and \citet{rudinger-etal-2018-gender}, we evaluate five freely available systems on the ablated data from \autoref{sec:datay}.
Third, we evaluate these systems on the dataset we created: Gender Inclusive Coreference (\newdataset).

\subsection{Cis-normativity in published NLP papers} \label{sec:published}
In our first study, we adapt the approach \citet{keyes18misgendering} took for analyzing the degree to which computer vision papers encoded trans-exclusive models of gender.
In particular, we began with a random sample of $\sim$150 papers from the ACL anthology that mention the word ``gender'' and coded them according to the following questions:
\begin{itemize}[leftmargin=*]
\item Does the paper discuss coreference resolution? 
\item Does the paper study English? 
\item \textbf{L.G}: Does the paper deal with linguistic gender (grammatical gender or gendered pronouns)? 
\item \textbf{S.G}: Does the paper deal with social gender? 
\item \textbf{L.G$\neq$S.G}: (If yes to L.G and S.G:) Does the paper distinguish linguistic from social gender? 
\item \textbf{S.G Binary}: (If yes to S.G:) Does the paper explicitly or implicitly assume that social gender is binary? 
\item \textbf{S.G Immutable}: (If yes to S.G:) Does the paper explicitly or implicitly assume social gender is immutable?
\item \textbf{They/Neo}: (If yes to S.G and to English:) Does the paper explicitly consider uses of definite singular ``they'' or neopronouns? 
\end{itemize}
The results of this coding are in \autoref{tab:papers} (the full annotation is in \autoref{s:annotation}).
We see out of the $22$ coreference papers analyzed, the vast majority conform to a ``folk'' theory of language:
\begin{itemize}[label=$\diamond$, leftmargin=*]
\item Only $5.5\%$ distinguish social from linguistic gender (despite it being relevant);
\item Only $5.6\%$ explicitly model gender as inclusive of non-binary identities;
\item No papers treat gender as anything other than completely immutable;\footnote{The most common ways in which papers implicitly assume that social gender is immutable is either 1) by relying on external knowledge bases that map names to ``gender''; or 2) by scraping a history of a user's social media posts or emails and assuming that their ``gender'' today matches the gender of that historical record.}
\item Only $7.1\%$ (one paper!) considers neopronouns and/or specific singular \pron{they}.
\end{itemize}
The situation for papers not specifically about coreference is similar (the majority of these papers are either purely linguistic papers about grammatical gender in languages other than English, or papers that do ``gender recognition'' of authors based on their writing;~\citet{may2019} discusses the (re)production of gender in automated gender recognition in NLP in much more detail). Overall, the situation more broadly is equally troubling, and generally also fails to escape from the folk theory of gender. In particular, none of the differences are significant at a $p=0.05$ level except for the first two questions, due to the small sample size (according to an $n-1$ chi-squared test).
The result is that although we do not know exactly what decisions are baked in to all systems, the vast majority in our study (including two papers by one of the authors \cite{daume05coref,daume15referring}) come with strong gender binary assumptions, and exist within a broader sphere of literature which erases non-binary and binary trans identities.


\begin{table}
  \centering\tablesize
  \begin{tabular}{rclcl}
    \toprule
    & \multicolumn{2}{c}{\textbf{All Papers}}    & \multicolumn{2}{c}{\textbf{Coref Papers}} \\
    \midrule
    \text{L.G?}             &  52.6\% & \textcolor{black!60!white}{\scriptsize (of 150)}
                                          &  95.4\% & \textcolor{black!60!white}{\scriptsize (of 22)} \\
    \text{S.G?}                 &  58.0\% & \textcolor{black!60!white}{\scriptsize (of 150)}
                                          &  86.3\% & \textcolor{black!60!white}{\scriptsize (of 22)} \\
    \text{L.G$\neq$S.G?}             &  11.1\% & \textcolor{black!60!white}{\scriptsize (of 27)}
                                          & ~~5.5\% & \textcolor{black!60!white}{\scriptsize (of 18)} \\
    \text{S.G Binary?}             &  92.8\% & \textcolor{black!60!white}{\scriptsize (of 84)}
                                          &  94.4\% & \textcolor{black!60!white}{\scriptsize (of 18)} \\
    \text{S.G Immutable?}              &  94.5\% & \textcolor{black!60!white}{\scriptsize (of 74)}
                                          & \!\!\!100.0\% & \textcolor{black!60!white}{\scriptsize (of 14)} \\
    \text{They/Neo?}    & ~~3.5\% & \textcolor{black!60!white}{\scriptsize (of 56)}
                                          & ~~7.1\% & \textcolor{black!60!white}{\scriptsize (of 14)} \\   
    \bottomrule
  \end{tabular}
  \caption{\label{tab:papers} Analysis of a corpus of 150 NLP papers that mention ``gender'' along the lines of what assumptions around gender are implicitly or explicitly made.}
\end{table}

\begin{figure}
    \centering
   
    {{\includegraphics[width=7cm]{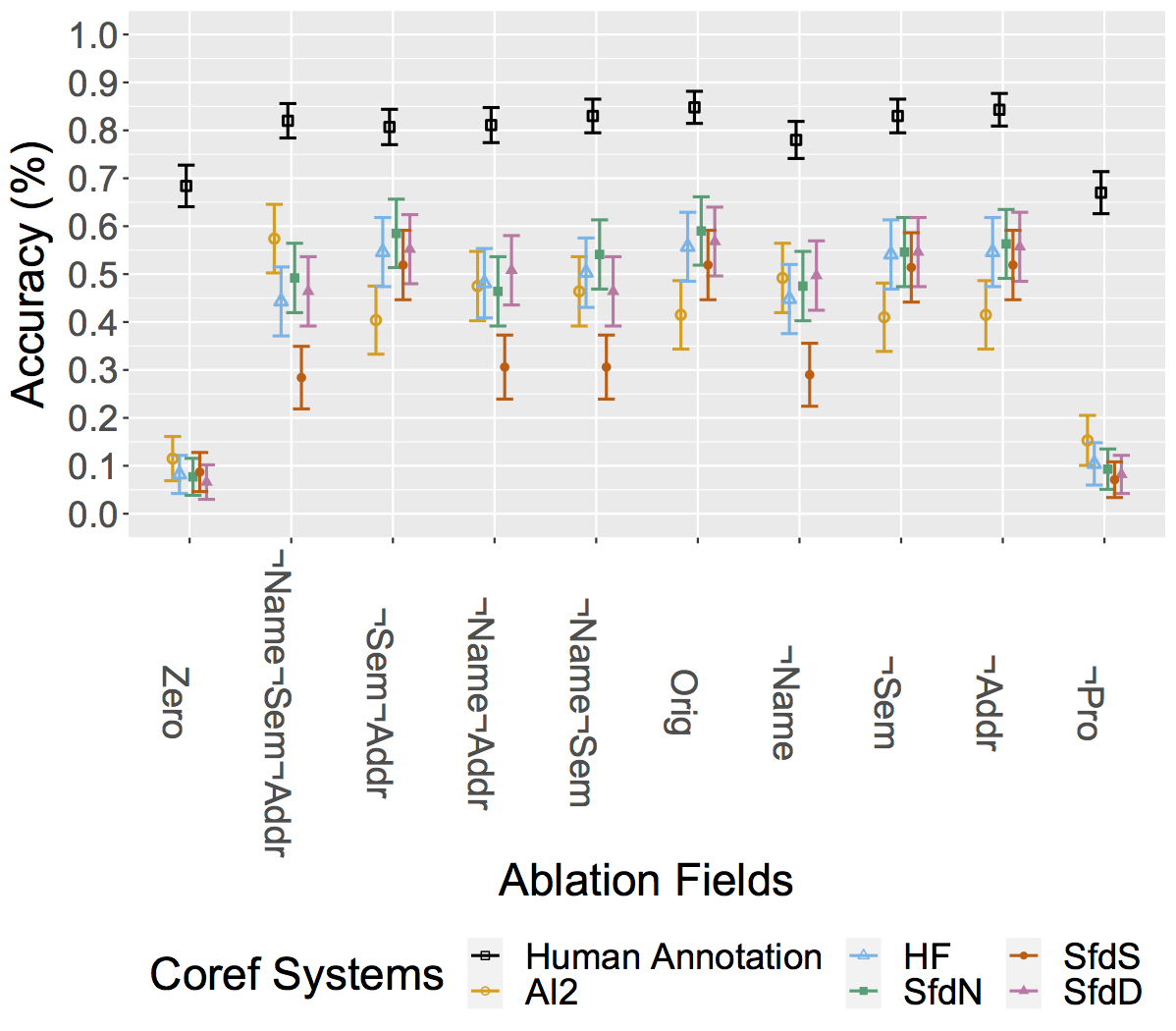} }}
    \caption{\label{fig:sys}  Coreference resolution systems results for the ablation study on \MAP dataset. The y-axis is the degree of \emph{accuracy} with 95\% significance intervals. }
    \label{Ablation results on systems}
\end{figure}

\subsection{System performance on MAP}
Next, we analyze the effect that our different ablation mechanisms have on existing coreference resolutions systems.
In particular, we run five coreference resolution systems on our ablated data: the AI2 system \citep[AI2; ][]{gardner2017allen}, hugging face \citep[HF; ][]{huggingface}, which is a neural system based on spacy, and the Stanford deterministic \citep[SfdD; ][]{raghunathan2010}, statistical \citep[SfdS; ][]{clark2015entity} and neural \citep[SfdN; ][]{clark2016deep} systems.
\autoref{fig:sys} shows the results. We can see that the system accuracies mostly follow the same pattern as human accuracy scores, though all are significantly lower than human results.
Accuracy scores for systems drop dramatically when we ablate out referential gender in pronouns. This reveals that those coreference resolution systems reply heavily on gender-based inferences. In terms of each systems, HF and SfdN systems have similar results and outperform other systems in most cases. SfdD accuracy drops significantly once names are ablated. 

These results echo and extend previous observations made by \citet{zhao-etal-2018-gender}, who focus on detecting stereotypes within occupations. They detect gender bias by checking if the system accuracies are the same for cases that can be resolved by syntactic cues and cases that cannot, with original data and reversed-gender data.
Similarly, \citet{rudinger-etal-2018-gender} focus on detecting stereotypes within occupations as well. They construct dataset without any gender cues other than stereotypes, and check how systems perform with different pronouns -- \pron{they}, \pron{she}, \pron{he}. Ideally, they should all perform the same because there is not any gender cues in the sentence. However, they find that systems do not work on ``they'' and perform better on ``he'' than ``she''.
Our analysis breaks this stereotyping down further to detect which aspects of gender signals are most leveraged by current systems.

\begin{table}
  \newcommand{\fixedmin}[1]{\rule{#1}{0pt}&\rule{#1}{0pt}&\rule{#1}{0pt}&\rule{#1}{0pt}\\[-\arraystretch\normalbaselineskip]}

  \centering\tablesize
  \begin{tabular}{lccc} \fixedmin{1.2cm}
    \toprule
    & \textbf{Precision} & \textbf{Recall}  & \textbf{F1}\\
    \midrule
    \textbf{AI2}     &     40.4\%  & \bf 29.2\%  &   33.9\%  \\
    \textbf{HF}      & \bf 68.8\%  &     22.3\%  &      33.6\%  \\
    \textbf{SfdD}    &     50.8\%  &     23.9\%  &      32.5\%  \\
    \textbf{SfdS}    &     59.8\%  &     24.1\%  &   \bf   34.3\%  \\
    \textbf{SfdN}    &     59.4\%  &     24.0\%  &      34.2\%  \\
    \bottomrule
  \end{tabular}
  \qquad
  \caption{\label{tab:newdatasetres} LEA scores on \newdataset (incorrect reference excluded) with various coreference resolution systems. Rows are different systems while columns are precision, recall, and F1 scores. When evaluate, we only count exact matches of pronouns and name entities. }
\end{table}

\subsection{System behavior on gender-inclusive data}\label{sec:gicoref}
Finally, in order to evaluate current coreference resolution models in gender inclusive contexts we introduce a new dataset, \newdataset.
Here we focused on \emph{naturally} occurring data, but sampled specifically to surface more gender-related phenomena than may be found in, say, the Wall Street Journal.

Our new \newdataset dataset consists of 95 documents from three types of sources: articles from English Wikipedia about people with non-binary gender identities, articles from LGBTQ periodicals, and fan-fiction stories from Archive Of Our Own (with the respective author's permission)\footnote{See \url{https://archiveofourown.org}; thanks to Os Keyes for this suggestion.}. These documents were each annotated by both of the authors and adjudicated.\footnote{We evaluate inter-annotator agreement by treating one annotation as gold standard and the other as system output and computing the LEA metric; the resulting F1-score is 92\%. During the adjudication process we found that most of the disagreement are due to one of the authors missing/overlooking mentions, and rarely due to true ``disagreement.''}
This data includes many examples of people who use pronouns other than \pron{she} or \pron{he}(the dataset contains 27\% \pron{he}, 20\% \pron{she}, 35\% \pron{they}, and 18\% neopronouns, people who are genderfluid and whose names or pronouns change through the article, people who are misgendered, and people in relationships that are not heteronormative. In addition, \emph{incorrect references}  (misgendering and deadnaming\footnote{According to \citet{healthline-deadnaming} deadnaming occurs when someone, intentionally or not, refers to a person who’s transgender by the name they used before they transitioned. }) are explicitly annotated.\footnote{Thanks to an anonymous reader of a draft version of this paper for this suggestion.} Two example annotated documents, one from Wikipedia, and one from Archive of Our Own, are provided in \autoref{ex:wiki} and \autoref{ex:ao3}.
%

We run the same systems as before on this dataset. \autoref{tab:newdatasetres} reports results according the standard coreference resolution evaluation metric LEA \cite{MoosaviLEA}. 
Since no systems are implemented to explicitly mark incorrect references, and no current evaluation metrics address this case, we perform the same evaluation twice. One with incorrect references included as regular references in the ground truth; and other with incorrect references excluded. 
Due to the limited number of incorrect references in the dataset, the difference of the results are not significant. Here we only report the latter.

The first observation is that there is still plenty room for coreference systems to improve; the best performing system achieves as F1 score of 34\%, but the Stanford neural system's F1 score on CoNLL-2012 test set reaches 60\% \citep{moosavi2020}.
Additionally, we can see system precision dominates recall.
This is likely partially due to poor recall of pronouns other than \pron{he} and \pron{she}. 
To analyze this, we compute the \emph{recall} of each system for finding referential pronouns at all, regardless of whether they are correctly linked to their antecedents.
We find that all systems achieve a recall of at least $95\%$ for binary pronouns, a recall of around $90\%$ on average for \pron{they}, and a recall of around a paltry $13\%$ for neopronouns (two systems---Stanford deterministic and Stanford neural---never identify any neopronouns at all).




%% file: discussion.tex
Our goal in this paper was to analyze how gender bias exist in coreference resolution annotations and models, with a particular focus on how it may fail to adequately process text involving binary and non-binary trans referents.
We thus created two datasets: \MAP and \newdataset .
Both datasets show significant gaps in system performance, but perhaps moreso, show that taking crowdworker judgments as ``gold standard'' can be problematic.
It may be the case that to truly build gender inclusive datasets and systems, we need to hire or consult experiential experts~\citep{patton2019annotating,You:Mag:Fri:19}.

Moreover, although we studied crowdworkers on Mechanical Turk (because they are often employed as annotators for NLP resources), if other populations are used for annotation, it becomes important to consider their positionality and how that may impact annotations. This echoes a related finding in annotation of hate-speech that annotator positionality matters \citep{olteanu2019social}.
More broadly, we found that trans-exclusionary assumptions around \emph{gender} in NLP papers is made commonly (and implicitly), a practice that we hope to see change in the future because it fundamentally limits the applicability of NLP systems.

The primary limitation of our study and analysis is that it is limited to English. This is particularly limiting because English lacks a grammatical gender system, and some extensions of our work to languages with grammatical gender are non-trivial. 
We also emphasize that while we endeavored to be inclusive, our own positionality has undoubtedly led to other biases.
One in particular is a largely Western bias, both in terms of what models of gender we use and also in terms of the data we annotated.
We have attempted to partially compensate for this bias by intentionally including documents with non-Western non-binary expressions of gender in the GICoref dataset\footnote{
    We endeavored to represent some non-Western gender identies that do not fall into the male/female binary, including people who identify as \emph{hijra} (Indian subcontinent), \emph{phuying} (Thailand, sometimes referred to as \emph{kathoey}), \emph{muxe} (Oaxaca), \emph{two-spirit} (Americas), \emph{fa'afafine} (Samoa) and \emph{m\=ah\=u} (Hawaii).}, but the dataset nonetheless remains Western-dominant.

Additionally, our ability to collect \emph{naturally occurring} data was limited because many sources simply do not yet permit (or have only recently permitted) the use of gender inclusive language in their articles.
This led us to counterfactual text manipulation, which, while useful, is essentially impossible to do flawlessly.
Moreover, our ability to evaluate coreference systems with data that includes \emph{incorrect references} was limited as well, because current systems do not mark any forms of misgendering or deadnaming explicitly, and current metrics do not take this into account.
Finally, because the social construct of gender is fundamentally contested, some of our results may apply only under some frameworks.

%

We hope this paper can serve as a roadmap for future studies.
In particular, the gender taxonomy we presented, while not novel, is (to our knowledge) previously unattested in discussions around gender bias in NLP systems; we hope future work in this area can draw on these ideas.
We also hope that developers of datasets or systems can use some of our analysis as inspiration for how one can attempt to measure---and then root out---different forms of bias in coreference resolution systems and NLP systems more broadly.


%% file: examples.tex

\section{Examples of Possible Bias in Data Annotation}\label{ex:biasexs}
Bias can enter coreference resolution datasets, which we use to train our systems, through annotation phase. Annotators may use linguistic notions to infer social gender. For instance, consider \lingref{john} below, in which an annotator is likely to determine that ``her'' refers to ``Mary'' and not ``John'' due to assumptions on likely ways that names may map to pronouns (or possibly by not considering that \pron{she} pronouns could refer to someone named ``John'').
While in \lingref{sue}, an annotator is likely to have difficulty making a determination because both ``Sue'' and ``Mary'' suggest ``her''.
In \lingref{liang}, an annotator lacking knowledge of name stereotypes on typical Chinese and Indian names (plus the fact that given names in Chinese --- especially when romanized  ---generally do not signal gender strongly), respectively, will likewise have difficulty. 

\lingex{john}{John and Mary visited \hlemph{her} mother.}
\lingex{sue}{Sue and Mary visited \hlemph{her} mother.}
\lingex{liang}{Liang and Aditya visited \hlemph{her} mother.}

\noindent
In all these cases, the plausible rough inference is that a reader takes a name, uses it to infer the social gender of the extra-linguistic referent.
Later the reader sees the \pron{she} pronoun, infers the referential gender of that pronoun, and checks to see if they match. 

An equivalent inference happens not just for names, but also for lexical gender references (both gendered nouns \lingref{lgen} and terms of address \lingref{lgen2}), grammatical gender references (in gender languages like Arabic \lingref{ggen}), and social gender references \lingref{sgen}.
The last of these (\lingref{sgen}) is the case in which the correct referent is likely to be least clear to most annotators, and also the case studied by \citet{rudinger-etal-2018-gender} and \citet{zhao-etal-2018-gender}.
\lingex{lgen}{My brother and niece visited \hlemph{her} mother.}
\lingex{lgen2}{Mr. Hashimoto and Mrs. Iwu visited \hlemph{her} mother.}
\lingex{ggen}{\parbox{\linewidth}{\includegraphics[width=5cm,trim={72 685 425 97},clip]{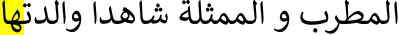}}
  \\\shortex{6}%
  {walidatu&\hlemph{-ha}  & shahidanaan & walidatuha                      & w   & almutarab}%
  {mother  &\hlemph{-her} & saw         & actor{\scriptsize[\feminine{}]} & and & singer{\scriptsize[\masculine{}]}}%
  {The singer{\scriptsize[\masculine{}]} and actor{\scriptsize[\feminine{}]} saw \hlemph{her} mother.}}
\lingex{sgen}{The nurse and the actor visited \hlemph{her} mother.}


\newpage
\section{Annotation of ACL Anthology Papers} \label{s:annotation}

\newcommand{\checkyes}{Y}
\newcommand{\checkno}{N}
\newcommand{\checkna}{-}

Below we list the complete set of annotations we did of the papers described in \autoref{sec:published}.
For each of the papers considered, we annotate the following items:
\begin{itemize}
  \item Coref:  Does the paper discuss coreference resolution? 
  \item L.G: Does the paper deal with linguistic gender (grammatical gender or gendered pronouns)? 
  \item S.G: Does the paper deal with social gender? 
  \item Eng: Does the paper study English? 
  \item L$\neq$G: (If yes to L.G and S.G:) Does the paper distinguish linguistic from social gender? 
  \item 0/1: (If yes to S.G:) Does the paper explicitly or implicitly assume that social gender is binary? 
  \item Imm: (If yes to S.G:) Does the paper explicitly or implicitly assume social gender is immutable?
  \item Neo: (If yes to S.G and to English:) Does the paper explicitly consider uses of definite singular ``they'' or neopronouns? 
\end{itemize}
For each of these, we mark with [\checkyes] if the answer is yes, [\checkno] if the answer is no, and [\checkna] if this question is not applicable (ie it doesn't pass the conditional checks).

\LTchunksize=200
\definecolor{lightgray}{gray}{0.9}
\let\oldlongtable\longtable
\let\endoldlongtable\endlongtable
\def\zza{\global\let\zz\zzb
\fullwidthcolor{lightgray}}%
\def\zzb{\global\let\zz\zza}
\def\fullwidthcolor#1{\color{#1}\leaders\vrule\hskip\textwidth\hskip-\textwidth\kern0pt}
\def\resetLTcolor{\global\let\zz\zza}
\LTleft0pt
\LTright0pt

\resetLTcolor
\begin{longtable}{@{\zz\extracolsep{\fill}} lcccccccc}
\toprule
\multicolumn{1}{@{\fullwidthcolor{white}\extracolsep{\fill}} l}{\bf Citation} & \bf Coref & \bf L.G & \bf S.G & \bf Eng & \bf L$\neq$S & \bf 0/1 & \bf Imm & \bf Neo \\
\midrule
\endhead
\citet{sidner-1981-focusing}                          & \checkyes & \checkyes & \checkyes & \checkyes & \checkno  & \checkna  & \checkna  & \checkna  \\
\citet{bainbridge-1985-montagovian}                   & \checkyes & \checkyes & \checkno  & \checkyes & \checkna  & \checkna  & \checkna  & \checkna  \\
\citet{kameyama-1986-property}                        & \checkyes & \checkyes & \checkyes & \checkyes & \checkno  & \checkyes & \checkyes & \checkno  \\
\citet{mellish-1988-implementing}                     & \checkno  & \checkyes & \checkno  & \checkyes & \checkna  & \checkna  & \checkna  & \checkna  \\
\citet{danlos-namer-1988-morphology}                  & \checkno  & \checkyes & \checkno  & \checkno  & \checkna  & \checkna  & \checkna  & \checkna  \\
\citet{yoshimoto-1988-identifying}                    & \checkno  & \checkyes & \checkno  & \checkno  & \checkna  & \checkna  & \checkna  & \checkna  \\
\citet{zock-etal-1988-language}                       & \checkno  & \checkyes & \checkno  & \checkno  & \checkna  & \checkna  & \checkna  & \checkna  \\
\citet{popowich-1989-tree}                            & \checkno  & \checkyes & \checkno  & \checkyes & \checkna  & \checkna  & \checkna  & \checkna  \\
\citet{mani-etal-1993-identifying}                    & \checkyes & \checkno  & \checkyes & \checkyes & \checkna  & \checkyes & \checkna  & \checkna  \\
\citet{narayanan-hashem-1993-abstract}                & \checkno  & \checkyes & \checkno  & \checkno  & \checkna  & \checkna  & \checkna  & \checkna  \\
\citet{soloman-wood-1994-learning}                    & \checkno  & \checkyes & \checkno  & \checkyes & \checkna  & \checkna  & \checkna  & \checkna  \\
\citet{quantz-1994-hpsg}                              & \checkno  & \checkyes & \checkno  & \checkyes & \checkna  & \checkna  & \checkna  & \checkna  \\
\citet{baker-etal-1994-research}                      & \checkna  & \checkna  & \checkna  & \checkna  & \checkna  & \checkna  & \checkna  & \checkna  \\
\citet{genthial-etal-1994-towards}                    & \checkno  & \checkyes & \checkno  & \checkno  & \checkna  & \checkna  & \checkna  & \checkna  \\
\citet{levinger-etal-1995-learning}                   & \checkno  & \checkyes & \checkno  & \checkno  & \checkna  & \checkna  & \checkna  & \checkna  \\
\citet{holan-etal-1997-prototype}                     & \checkno  & \checkyes & \checkno  & \checkno  & \checkna  & \checkna  & \checkna  & \checkna  \\
\citet{dorna-etal-1998-syntactic}                     & \checkno  & \checkno  & \checkno  & \checkyes & \checkna  & \checkna  & \checkna  & \checkna  \\
\citet{harabagiu-maiorano-1999-knowledge}             & \checkyes & \checkyes & \checkyes & \checkyes & \checkno  & \checkyes & \checkyes & \checkno  \\
\citet{avgustinova-uszkoreit-2000-ontology}           & \checkno  & \checkyes & \checkno  & \checkno  & \checkna  & \checkna  & \checkna  & \checkna  \\
\citet{channarukul-etal-2000-enriching}               & \checkno  & \checkyes & \checkno  & \checkyes & \checkna  & \checkna  & \checkna  & \checkna  \\
\citet{abuleil-etal-2002-acquisition}                 & \checkno  & \checkyes & \checkno  & \checkno  & \checkna  & \checkna  & \checkna  & \checkna  \\
\citet{cucerzan-yarowsky-2003-minimally}              & \checkno  & \checkyes & \checkno  & \checkno  & \checkna  & \checkna  & \checkna  & \checkna  \\
\citet{pakhomov-etal-2003-identification}             & \checkno  & \checkno  & \checkyes & \checkyes & \checkna  & \checkna  & \checkna  & \checkna  \\
\citet{tadic-fulgosi-2003-building}                   & \checkno  & \checkyes & \checkno  & \checkno  & \checkna  & \checkna  & \checkna  & \checkna  \\
\citet{debowski-2003-reconfigurable}                  & \checkno  & \checkyes & \checkno  & \checkno  & \checkna  & \checkna  & \checkna  & \checkna  \\
\citet{navarretta-2004-algorithm}                     & \checkyes & \checkyes & \checkyes & \checkno  & \checkno  & \checkyes & \checkyes & \checkna  \\
\citet{carl-etal-2004-controlling}                    & \checkyes & \checkyes & \checkyes & \checkno  & \checkno  & \checkyes & \checkyes & \checkna  \\
\citet{mota-etal-2004-multiword}                      & \checkno  & \checkyes & \checkno  & \checkyes & \checkna  & \checkna  & \checkna  & \checkna  \\
\citet{eisner-karakos-2005-bootstrapping}             & \checkno  & \checkyes & \checkno  & \checkyes & \checkna  & \checkna  & \checkna  & \checkna  \\
\citet{boulis-ostendorf-2005-quantitative}            & \checkno  & \checkno  & \checkyes & \checkyes & \checkna  & \checkyes & \checkyes & \checkno  \\
\citet{smith-etal-2005-context}                       & \checkno  & \checkyes & \checkno  & \checkno  & \checkna  & \checkna  & \checkna  & \checkna  \\
\citet{bergsma-lin-2006-bootstrapping}                & \checkyes & \checkyes & \checkyes & \checkyes & \checkno  & \checkyes & \checkyes & \checkno  \\
\citet{vogt-andre-2006-improving}                     & \checkno  & \checkno  & \checkyes & \checkno  & \checkna  & \checkyes & \checkyes & \checkna  \\
\citet{quirk-corston-oliver-2006-impact}              & \checkno  & \checkyes & \checkno  & \checkyes & \checkna  & \checkna  & \checkna  & \checkna  \\
\citet{dada-2007-implementation}                      & \checkno  & \checkyes & \checkno  & \checkno  & \checkna  & \checkna  & \checkna  & \checkna  \\
\citet{streiter-etal-2007-tombstones}                 & \checkno  & \checkno  & \checkyes & \checkno  & \checkna  & \checkna  & \checkna  & \checkna  \\
\citet{jing-etal-2007-extracting}                     & \checkyes & \checkyes & \checkyes & \checkyes & \checkno  & \checkyes & \checkna  & \checkno  \\
\citet{badr-etal-2008-segmentation}                   & \checkno  & \checkyes & \checkno  & \checkno  & \checkna  & \checkna  & \checkna  & \checkna  \\
\citet{marchal-etal-2008-mdl}                         & \checkno  & \checkyes & \checkno  & \checkno  & \checkna  & \checkna  & \checkna  & \checkna  \\
\citet{van-peursen-2009-establish}                    & \checkno  & \checkyes & \checkno  & \checkno  & \checkna  & \checkna  & \checkna  & \checkna  \\
\citet{badr-etal-2009-syntactic}                      & \checkno  & \checkyes & \checkno  & \checkno  & \checkna  & \checkna  & \checkna  & \checkna  \\
\citet{garera-yarowsky-2009-modeling}                 & \checkno  & \checkyes & \checkyes & \checkyes & \checkno  & \checkyes & \checkyes & \checkno  \\
\citet{bergsma-etal-2009-glen}                        & \checkyes & \checkyes & \checkyes & \checkyes & \checkno  & \checkyes & \checkyes & \checkno  \\
\citet{nastase-popescu-2009-whats}                    & \checkno  & \checkyes & \checkno  & \checkno  & \checkna  & \checkna  & \checkna  & \checkna  \\
\citet{nanba-etal-2009-automatic}                     & \checkno  & \checkno  & \checkno  & \checkyes & \checkna  & \checkna  & \checkna  & \checkna  \\
\citet{robaldo-di-carlo-2009-disambiguating}          & \checkno  & \checkno  & \checkno  & \checkyes & \checkna  & \checkna  & \checkna  & \checkna  \\
\citet{mukherjee-liu-2010-improving}                  & \checkno  & \checkno  & \checkyes & \checkyes & \checkna  & \checkyes & \checkyes & \checkna  \\
\citet{ng-2010-supervised}                            & \checkyes & \checkyes & \checkyes & \checkyes & \checkno  & \checkyes & \checkyes & \checkno  \\
\citet{burkhardt-etal-2010-database}                  & \checkno  & \checkno  & \checkyes & \checkno  & \checkna  & \checkyes & \checkyes & \checkna  \\
\citet{marton-etal-2010-improving}                    & \checkno  & \checkyes & \checkno  & \checkno  & \checkna  & \checkna  & \checkna  & \checkna  \\
\citet{le-nagard-koehn-2010-aiding}                   & \checkyes & \checkyes & \checkyes & \checkyes & \checkno  & \checkyes & \checkyes & \checkno  \\
\citet{rojas-barahona-etal-2011-using}                & \checkno  & \checkyes & \checkno  & \checkno  & \checkna  & \checkna  & \checkna  & \checkna  \\
\citet{mukund-etal-2011-using}                        & \checkno  & \checkyes & \checkno  & \checkno  & \checkna  & \checkna  & \checkna  & \checkna  \\
\citet{sarawgi-etal-2011-gender}                      & \checkno  & \checkno  & \checkyes & \checkyes & \checkna  & \checkyes & \checkyes & \checkno  \\
\citet{li-etal-2011-pronoun}                          & \checkyes & \checkyes & \checkyes & \checkyes & \checkno  & \checkyes & \checkyes & \checkno  \\
\citet{burger-etal-2011-discriminating}               & \checkno  & \checkno  & \checkyes & \checkyes & \checkna  & \checkyes & \checkyes & \checkno  \\
\citet{mohammad-yang-2011-tracking}                   & \checkno  & \checkno  & \checkyes & \checkyes & \checkna  & \checkyes & \checkyes & \checkno  \\
\citet{sapena-etal-2011-relaxcor}                     & \checkyes & \checkyes & \checkyes & \checkyes & \checkno  & \checkyes & \checkyes & \checkno  \\
\citet{charton-gagnon-2011-poly}                      & \checkyes & \checkyes & \checkyes & \checkyes & \checkno  & \checkyes & \checkyes & \checkno  \\
\citet{alkuhlani-habash-2011-corpus}                  & \checkno  & \checkyes & \checkno  & \checkno  & \checkna  & \checkna  & \checkna  & \checkna  \\
\citet{marecek-etal-2011-two}                         & \checkno  & \checkyes & \checkno  & \checkno  & \checkna  & \checkna  & \checkna  & \checkna  \\
\citet{lopez-ludena-etal-2011-source}                 & \checkno  & \checkyes & \checkno  & \checkno  & \checkna  & \checkna  & \checkna  & \checkna  \\
\citet{declerck-etal-2012-ontology}                   & \checkyes & \checkyes & \checkno  & \checkyes & \checkna  & \checkna  & \checkna  & \checkna  \\
\citet{bergsma-etal-2012-stylometric}                 & \checkno  & \checkno  & \checkyes & \checkyes & \checkna  & \checkyes & \checkyes & \checkno  \\
\citet{alkuhlani-habash-2012-identifying}             & \checkno  & \checkyes & \checkno  & \checkno  & \checkna  & \checkna  & \checkna  & \checkna  \\
\citet{filippova-2012-user}                           & \checkno  & \checkno  & \checkyes & \checkyes & \checkna  & \checkyes & \checkna  & \checkna  \\
\citet{dinu-etal-2012-romanian}                       & \checkno  & \checkyes & \checkno  & \checkno  & \checkna  & \checkna  & \checkna  & \checkna  \\
\citet{el-kholy-habash-2012-rich}                     & \checkno  & \checkyes & \checkno  & \checkno  & \checkna  & \checkna  & \checkna  & \checkna  \\
\citet{yu-2012-function}                              & \checkno  & \checkno  & \checkno  & \checkno  & \checkna  & \checkna  & \checkna  & \checkna  \\
\citet{guillou-2012-improving}                        & \checkyes & \checkyes & \checkyes & \checkyes & \checkyes & \checkyes & \checkna  & \checkna  \\
\citet{vogel-jurafsky-2012-said}                      & \checkno  & \checkno  & \checkyes & \checkyes & \checkna  & \checkyes & \checkyes & \checkno  \\
\citet{goldberg-elhadad-2013-word}                    & \checkno  & \checkyes & \checkno  & \checkno  & \checkna  & \checkna  & \checkna  & \checkna  \\
\citet{marton-etal-2013-dependency}                   & \checkno  & \checkyes & \checkno  & \checkno  & \checkna  & \checkna  & \checkna  & \checkna  \\
\citet{weller-etal-2013-using}                        & \checkno  & \checkyes & \checkno  & \checkyes & \checkna  & \checkna  & \checkna  & \checkna  \\
\citet{ciot-etal-2013-gender}                         & \checkno  & \checkno  & \checkyes & \checkno  & \checkna  & \checkyes & \checkyes & \checkna  \\
\citet{volkova-etal-2013-exploring}                   & \checkno  & \checkno  & \checkyes & \checkyes & \checkna  & \checkyes & \checkyes & \checkno  \\
\citet{levitan-2013-entrainment}                      & \checkno  & \checkno  & \checkyes & \checkyes & \checkna  & \checkno  & \checkno  & \checkno  \\
\citet{bojar-etal-2013-chimera}                       & \checkno  & \checkyes & \checkno  & \checkno  & \checkna  & \checkna  & \checkna  & \checkna  \\
\citet{glavas-etal-2013-aspect}                       & \checkno  & \checkyes & \checkno  & \checkno  & \checkna  & \checkna  & \checkna  & \checkna  \\
\citet{liu-etal-2013-pal}                             & \checkno  & \checkno  & \checkno  & \checkno  & \checkna  & \checkna  & \checkna  & \checkna  \\
\citet{kestemont-2014-function}                       & \checkno  & \checkno  & \checkno  & \checkyes & \checkna  & \checkna  & \checkna  & \checkna  \\
\citet{novak-zabokrtsky-2014-cross}                   & \checkyes & \checkyes & \checkno  & \checkyes & \checkna  & \checkna  & \checkna  & \checkna  \\
\citet{babych-etal-2014-deriving}                     & \checkno  & \checkyes & \checkno  & \checkno  & \checkna  & \checkna  & \checkna  & \checkna  \\
\citet{soler-company-wanner-2014-use}                 & \checkno  & \checkno  & \checkyes & \checkyes & \checkna  & \checkyes & \checkyes & \checkno  \\
\citet{chen-ng-2014-chinese}                          & \checkyes & \checkyes & \checkyes & \checkyes & \checkno  & \checkyes & \checkyes & \checkno  \\
\citet{sap-etal-2014-developing}                      & \checkno  & \checkno  & \checkyes & \checkyes & \checkna  & \checkyes & \checkyes & \checkna  \\
\citet{nguyen-etal-2014-gender}                       & \checkno  & \checkno  & \checkyes & \checkyes & \checkna  & \checkyes & \checkyes & \checkno  \\
\citet{prabhakaran-etal-2014-gender}                  & \checkno  & \checkno  & \checkyes & \checkyes & \checkna  & \checkyes & \checkyes & \checkno  \\
\citet{sidorov-etal-2014-comparison}                  & \checkno  & \checkno  & \checkyes & \checkyes & \checkna  & \checkyes & \checkyes & \checkno  \\
\citet{darwish-etal-2014-using}                       & \checkno  & \checkyes & \checkno  & \checkno  & \checkna  & \checkna  & \checkna  & \checkna  \\
\citet{ahmed-khan-2014-automatic}                     & \checkno  & \checkyes & \checkno  & \checkno  & \checkna  & \checkna  & \checkna  & \checkna  \\
\citet{nguyen-etal-2014-tweetgenie}                   & \checkno  & \checkno  & \checkyes & \checkno  & \checkna  & \checkyes & \checkyes & \checkna  \\
\citet{stewart-2014-now}                              & \checkno  & \checkno  & \checkyes & \checkyes & \checkna  & \checkyes & \checkyes & \checkna  \\
\citet{matthews-etal-2014-cmu}                        & \checkno  & \checkyes & \checkno  & \checkno  & \checkna  & \checkna  & \checkna  & \checkna  \\
\citet{vaidya-etal-2014-light}                        & \checkno  & \checkyes & \checkno  & \checkno  & \checkna  & \checkna  & \checkna  & \checkna  \\
\citet{kokkinakis-etal-2015-gender}                   & \checkno  & \checkyes & \checkyes & \checkno  & \checkno  & \checkyes & \checkna  & \checkna  \\
\citet{johannsen-etal-2015-cross}                     & \checkno  & \checkno  & \checkyes & \checkyes & \checkna  & \checkyes & \checkyes & \checkna  \\
\citet{schwartz-etal-2015-extracting}                 & \checkno  & \checkno  & \checkno  & \checkyes & \checkna  & \checkna  & \checkna  & \checkna  \\
\citet{hovy-2015-demographic}                         & \checkno  & \checkno  & \checkyes & \checkyes & \checkna  & \checkyes & \checkyes & \checkno  \\
\citet{agarwal-etal-2015-key}                         & \checkno  & \checkyes & \checkyes & \checkyes & \checkno  & \checkyes & \checkyes & \checkno  \\
\citet{preotiuc-pietro-etal-2015-role}                & \checkno  & \checkno  & \checkyes & \checkyes & \checkno  & \checkyes & \checkyes & \checkna  \\
\citet{ramakrishna-etal-2015-quantitative}            & \checkno  & \checkyes & \checkyes & \checkyes & \checkno  & \checkyes & \checkyes & \checkno  \\
\citet{taniguchi-etal-2015-weighted}                  & \checkno  & \checkno  & \checkyes & \checkyes & \checkna  & \checkno  & \checkyes & \checkno  \\
\citet{schofield-mehr-2016-gender}                    & \checkno  & \checkno  & \checkyes & \checkyes & \checkna  & \checkyes & \checkyes & \checkno  \\
\citet{levitan-etal-2016-identifying}                 & \checkno  & \checkno  & \checkyes & \checkyes & \checkna  & \checkyes & \checkyes & \checkno  \\
\citet{flekova-etal-2016-analyzing}                   & \checkno  & \checkno  & \checkyes & \checkyes & \checkna  & \checkyes & \checkyes & \checkno  \\
\citet{tran-ostendorf-2016-characterizing}            & \checkno  & \checkno  & \checkno  & \checkyes & \checkna  & \checkna  & \checkna  & \checkna  \\
\citet{qian-etal-2016-investigating}                  & \checkno  & \checkyes & \checkno  & \checkyes & \checkna  & \checkna  & \checkna  & \checkna  \\
\citet{li-etal-2016-semi}                             & \checkno  & \checkno  & \checkyes & \checkyes & \checkna  & \checkyes & \checkyes & \checkno  \\
\citet{zhang-etal-2016-user}                          & \checkno  & \checkno  & \checkyes & \checkyes & \checkna  & \checkyes & \checkyes & \checkno  \\
\citet{garimella-mihalcea-2016-zooming}               & \checkno  & \checkno  & \checkyes & \checkyes & \checkna  & \checkyes & \checkyes & \checkno  \\
\citet{reddy-knight-2016-obfuscating}                 & \checkno  & \checkno  & \checkyes & \checkyes & \checkna  & \checkyes & \checkyes & \checkno  \\
\citet{li-dickinson-2017-gender}                      & \checkno  & \checkno  & \checkyes & \checkno  & \checkna  & \checkyes & \checkyes & \checkna  \\
\citet{perez-estruch-etal-2017-learning}              & \checkno  & \checkno  & \checkyes & \checkyes & \checkna  & \checkyes & \checkyes & \checkno  \\
\citet{perez-rosas-etal-2017-identity}                & \checkno  & \checkno  & \checkyes & \checkyes & \checkna  & \checkyes & \checkyes & \checkno  \\
\citet{rabinovich-etal-2017-personalized}             & \checkno  & \checkno  & \checkyes & \checkno  & \checkna  & \checkyes & \checkyes & \checkna  \\
\citet{costa-jussa-2017-catalan}                      & \checkno  & \checkyes & \checkno  & \checkno  & \checkna  & \checkna  & \checkna  & \checkna  \\
\citet{sap-etal-2017-connotation}                     & \checkno  & \checkno  & \checkyes & \checkyes & \checkna  & \checkyes & \checkna  & \checkna  \\
\citet{zhao-etal-2017-men}                            & \checkno  & \checkno  & \checkyes & \checkyes & \checkna  & \checkyes & \checkyes & \checkno  \\
\citet{mandravickaite-krilavicius-2017-stylometric}   & \checkno  & \checkno  & \checkyes & \checkyes & \checkna  & \checkyes & \checkyes & \checkno  \\
\citet{verhoeven-etal-2017-gender}                    & \checkno  & \checkno  & \checkyes & \checkyes & \checkna  & \checkyes & \checkyes & \checkno  \\
\citet{larson-2017-gender}                            & \checkno  & \checkyes & \checkyes & \checkyes & \checkyes & \checkno  & \checkno  & \checkyes \\
\citet{koolen-van-cranenburgh-2017-stereotypes}       & \checkno  & \checkno  & \checkyes & \checkno  & \checkna  & \checkno  & \checkyes & \checkna  \\
\citet{tatman-2017-gender}                            & \checkno  & \checkno  & \checkyes & \checkyes & \checkna  & \checkyes & \checkyes & \checkno  \\
\citet{soler-company-wanner-2017-relevance}           & \checkno  & \checkno  & \checkyes & \checkyes & \checkna  & \checkyes & \checkyes & \checkno  \\
\citet{ljubesic-etal-2017-language}                   & \checkno  & \checkno  & \checkyes & \checkno  & \checkna  & \checkyes & \checkyes & \checkna  \\
\citet{litvinova-etal-2017-deception}                 & \checkno  & \checkno  & \checkyes & \checkno  & \checkna  & \checkyes & \checkyes & \checkna  \\
\citet{mohammad-etal-2018-semeval}                    & \checkno  & \checkno  & \checkyes & \checkyes & \checkna  & \checkyes & \checkna  & \checkna  \\
\citet{wang-jurgens-2018-going}                       & \checkno  & \checkyes & \checkyes & \checkyes & \checkyes & \checkno  & \checkno  & \checkno  \\
\citet{kraus-etal-2018-effects}                       & \checkno  & \checkno  & \checkyes & \checkyes & \checkna  & \checkyes & \checkna  & \checkna  \\
\citet{martinc-pollak-2018-reusable}                  & \checkno  & \checkno  & \checkyes & \checkyes & \checkna  & \checkyes & \checkyes & \checkno  \\
\citet{chan-fyshe-2018-social}                        & \checkno  & \checkno  & \checkyes & \checkyes & \checkna  & \checkyes & \checkyes & \checkno  \\
\citet{durmus-cardie-2018-understanding}              & \checkno  & \checkno  & \checkno  & \checkyes & \checkna  & \checkna  & \checkna  & \checkna  \\
\citet{zaghouani-charfi-2018-arap}                    & \checkno  & \checkyes & \checkyes & \checkno  & \checkno  & \checkyes & \checkyes & \checkna  \\
\citet{plank-2018-predicting}                         & \checkno  & \checkno  & \checkyes & \checkyes & \checkna  & \checkyes & \checkyes & \checkno  \\
\citet{wood-doughty-etal-2018-predicting}             & \checkno  & \checkno  & \checkyes & \checkyes & \checkna  & \checkyes & \checkyes & \checkno  \\
\citet{moorthy-etal-2018-nike}                        & \checkno  & \checkno  & \checkyes & \checkyes & \checkna  & \checkyes & \checkna  & \checkna  \\
\citet{levitan-etal-2018-linguistic}                  & \checkno  & \checkno  & \checkyes & \checkyes & \checkna  & \checkyes & \checkyes & \checkno  \\
\citet{webster-etal-2018-mind}                        & \checkyes & \checkyes & \checkyes & \checkyes & \checkno  & \checkyes & \checkyes & \checkno  \\
\citet{park-etal-2018-reducing}                       & \checkno  & \checkyes & \checkyes & \checkyes & \checkno  & \checkyes & \checkyes & \checkno  \\
\citet{vanmassenhove-etal-2018-getting}               & \checkno  & \checkyes & \checkyes & \checkno  & \checkno  & \checkyes & \checkyes & \checkna  \\
\citet{kleinberg-etal-2018-identifying}               & \checkno  & \checkno  & \checkyes & \checkyes & \checkna  & \checkyes & \checkyes & \checkno  \\
\citet{zhao-etal-2018-learning}                       & \checkno  & \checkno  & \checkyes & \checkyes & \checkna  & \checkyes & \checkyes & \checkno  \\
\citet{balusu-etal-2018-stylistic}                    & \checkno  & \checkno  & \checkno  & \checkyes & \checkna  & \checkna  & \checkna  & \checkna  \\
\citet{rudinger-etal-2018-gender}                     & \checkyes & \checkyes & \checkyes & \checkyes & \checkno  & \checkno  & \checkna  & \checkyes \\
\citet{zhao-etal-2018-gender}                         & \checkyes & \checkyes & \checkyes & \checkyes & \checkno  & \checkyes & \checkyes & \checkno  \\
\citet{kiritchenko-mohammad-2018-examining}           & \checkna  & \checkna  & \checkna  & \checkna  & \checkna  & \checkna  & \checkna  & \checkna  \\
\citet{barbieri-camacho-collados-2018-gender}         & \checkno  & \checkno  & \checkyes & \checkyes & \checkna  & \checkyes & \checkno  & \checkna  \\
\citet{van-der-goot-etal-2018-bleaching}              & \checkno  & \checkno  & \checkyes & \checkno  & \checkna  & \checkyes & \checkyes & \checkna  \\
\citet{karlekar-etal-2018-detecting}                  & \checkno  & \checkno  & \checkyes & \checkyes & \checkna  & \checkyes & \checkyes & \checkno  \\
\citet{de-gibert-etal-2018-hate}                      & \checkno  & \checkno  & \checkno  & \checkyes & \checkna  & \checkna  & \checkna  & \checkna  \\
\citet{mickus-etal-2019-distributional}               & \checkno  & \checkyes & \checkno  & \checkno  & \checkna  & \checkna  & \checkna  & \checkna  \\
\bottomrule
\end{longtable}

\newpage

\section{Example GICoref Document from Wikipedia: Dana Zzyym} \label{ex:wiki}

\setlength{\fboxsep}{0.5pt}
\newcommand{\nmystrut}{\rule[-.15\baselineskip]{0pt}{0.7\baselineskip}}
\newcommand{\corefbox}[3]{\fcolorbox{#2!80!white}{#2!20!white}{\nmystrut{#3}}$_{\textcolor{#2!80!white}{\textsf{#1}}}$}
\newcommand{\corefboxa}[1]{\corefbox{A}{red}{#1}}
\newcommand{\corefboxb}[1]{\corefbox{B}{blue}{#1}}
\newcommand{\corefboxc}[1]{\corefbox{C}{orange}{#1}}
\newcommand{\corefboxd}[1]{\corefbox{D}{purple}{#1}}
\newcommand{\corefboxe}[1]{\corefbox{E}{cyan}{#1}}
\newcommand{\corefboxf}[1]{\corefbox{F}{brown}{#1}}
\newcommand{\corefboxg}[1]{\corefbox{G}{yellow}{#1}}
\newcommand{\corefboxh}[1]{\corefbox{H}{green}{#1}}
\newcommand{\corefboxi}[1]{\corefbox{I}{pink}{#1}}
\newcommand{\corefboxj}[1]{\corefbox{J}{grey}{#1}}
\newcommand{\corefboxk}[1]{\corefbox{K}{teal}{#1}}
\newcommand{\corefboxl}[1]{\corefbox{L}{violet}{#1}}

\textbf{[[Source: \url{https://en.wikipedia.org/wiki/Dana_Zzyym}]]}

~

\begin{footnotesize} \sffamily
\corefboxa{Dana Alix Zzyym} is an Intersex activist and former sailor who was the first military veteran in the United States to seek a non - binary gender U.S. passport , in a lawsuit \corefboxa{Zzyym} v. \corefboxc{Pompeo} .

~

\noindent
Early life

\corefboxa{Zzyym} has expressed that \corefboxa{their} childhood as a military brat made it out of the question for \corefboxa{them} to be associated with the queer community as a youth due to the prevalence of homophobia in the armed forces .
\corefboxb{\corefboxa{Their} parents} hid \corefboxa{Zzyym} 's status as intersex from \corefboxa{them} and \corefboxa{Zzyym} discovered \corefboxa{their} identity and the surgeries \corefboxb{\corefboxa{their} parents} had approved for \corefboxa{them} by \corefboxb{themselves} after \corefboxa{their} Navy service .
In 1978 , \corefboxa{Zzyym} joined the Navy as a machinist 's mate .

~

\noindent
Activism

\corefboxa{Zzyym} has been an avid supporter of the Intersex Campaign for Equality .

~

\noindent
Legal case

\corefboxa{Zzyym} is the first veteran to seek a non - binary gender U.S. passport .
In light of the State Department 's continuing refusal to recognize an appropriate gender marker , on June 27 , 2017 a federal court granted Lambda Legal 's motion to reopen the case .
On September 19 , 2018 , the United States District Court for the District of Colorado enjoined the U.S. Department of State from relying upon its binary - only gender marker policy to withhold the requested passport .
\end{footnotesize}

\newpage

\section{Example GICoref Document from AO3: Scar Tissue} \label{ex:ao3}

\textbf{[[Source: \url{https://archiveofourown.org/works/14476524}]]}

\textbf{[[Author: cornheck]]}

~

\begin{footnotesize} \sffamily
  Despite dreading \corefboxa{their} first true series of final exams , \corefboxa{Crona} 's relieved to have a particularly absorbative memory , lucky to recall all the material \corefboxa{they} 'd been required to catch up on .
  Half a semester of attendance , a whole year of course content .

  The only true moment of discomfort came when \corefboxa{they} 'd arrived at the essay portion .
  Thankful it was easy enough to answer , however , \corefboxa{their} subtle eye - roll stemmed entirely from just how much writing it asked of \corefboxa{them} , hands already beginning to ache at the thought of scrawling out two pages on the origins , history , and importance of partnered and grouped soul resonance .

  By the end of it all , \corefboxa{their} neck , wrist , back , and ribs ached from the strain of \corefboxa{their} typical , hunched posture -- a habit \corefboxa{they} defaulted to , and \corefboxb{Miss Marie} silently wished \corefboxa{they} 'd be more mindful of .
  It was a relief , at least to \corefboxa{them} , not to be the last one out of the lecture hall .
  Booklet turned in , \corefboxa{they} left the room as quietly as possible and lingered just outside , an air of hesitance settling upon \corefboxa{them} as \corefboxa{they} considered what to do now that , it seemed , everything was over with .
  No more class , no more lessons , just ... students on break from their studies for the season .

  `` Kind of a breeze , was n't it ? ''
  \corefboxc{Evans} ' voice echoes in the arched hall and \corefboxa{Crona} 's shoulders jump , \corefboxa{their} frame still a tense and anxious mess .

  `` Oh , '' \corefboxa{they} sigh ,
  `` \corefboxa{I} ... \corefboxa{I} suppose so .
  It was n't ... necessarily hard . ''
  \corefboxa{Crona} answers , putting forth a vaguely forced smile .

  Smiling with the assumed purpose of making \corefboxc{Soul} comfortable with the interaction .
  A defense mechanism .

  `` \corefboxa{I} - \corefboxa{I} guess , for a final , it was easier than \corefboxa{I} expected ... everyone ... made it sound like it 'd be difficult . ''

  `` If by everyone , \corefboxa{you} mean \corefboxd{Black Star} , then yeah , '' \corefboxc{Soul} chuckles , `` \corefboxd{he} does n't really do well on ` em ... bad test - taker . ''

  `` Ah , '' \corefboxa{their} facade falls just in time to be replaced by a much more genuine grin .

  Of the little \corefboxa{they} 'd spent talking to \corefboxd{Black Star} , \corefboxd{he} certainly had confidence and skill enough to make up for the lost exam points given \corefboxd{his} performance in every other grading category .

  `` That ... makes sense . ''

  `` \corefboxe{Maka} 's always the first one done when it comes to this stuff , \corefboxe{she} practically studies in \corefboxe{her} sleep .
  \corefboxc{I} 'm convinced \corefboxe{she} must be practicing clairvoyance the way \corefboxe{she} burns through essay questions , '' \corefboxc{Soul} laughs , turning to \corefboxa{the meek teen} who gives \corefboxc{him} a simple nod in response .

  Determined not to let an impending awkward silence fall between \corefboxf{them} , \corefboxc{Soul} pipes up again , `` So , are \corefboxa{you} staying here for break ? ''

  `` Ye - well , \corefboxa{I} ... \corefboxa{I} think so , '' \corefboxa{they} begin , stuttering , but encouraged to continue by a cock of \corefboxc{Soul} 's head ; a social cue even \corefboxa{they} could read , `` \corefboxg{\corefboxh{The professor} ... and \corefboxb{Miss Marie}} asked if \corefboxa{I} 'd like to come and stay with \corefboxg{them} for the time being . ''

  `` Oh , huh , \corefboxg{\corefboxh{Stein} and \corefboxb{Marie}} ?
  Nice , '' \corefboxc{his} brows lift , clearly some varying degree of happy for \corefboxa{the other} .

  The optimism is short - lived , observing as \corefboxa{Crona} 's expression falls back to its characteristic expressionless gaze .

  `` It seems like \corefboxa{you} 've got a good thing going with \corefboxg{those two} . ''

  `` \corefboxa{I} have n't decided , yet , if \corefboxa{I} should accept the invitation , '' \corefboxa{they} shift a bit where \corefboxa{they} stand .

  Never having been the best at reassuring others , even \corefboxa{\corefboxc{his} own meister} , \corefboxc{Soul} kept \corefboxc{his} mouth shut to avoid stuttering while \corefboxc{he} searched for the right words a web of thoughts .

  `` \corefboxa{Y '} know , \corefboxc{I} think it 's less of an invitation and more of an extended welcome . ''

  \corefboxa{The other} raises \corefboxa{their} head , taken aback , `` Oh , '' \corefboxa{Crona} mutters , in a poignant tone , `` \corefboxa{I} ... never considered something like that . ''
  
  \corefboxc{Soul} does n't leave much wiggle room for \corefboxa{their} mood to fall any further ( nothing past a flat - lipped frown ) , `` \corefboxg{They} 'd probably love to have \corefboxa{you} , \corefboxc{I} bet \corefboxg{they} drive each other nuts sometimes all by \corefboxg{themselves} . ''
  
  Though \corefboxc{Evans} wo n't admit it , \corefboxc{he} knows it 's all too likely \corefboxh{Stein} might actually put some more effort into taking care of \corefboxh{himself} if \corefboxh{he} had someone else besides \corefboxb{Marie} to look after .

  `` \corefboxa{I} - \corefboxa{I} see , '' \corefboxa{they} exhale with a nod , giving \corefboxc{Soul} a hint of affirmation that \corefboxc{he} 'd done something to boost \corefboxa{the kid} 's confidence .
  
  `` \corefboxc{I} mean , it 's got ta be lonely not to mention boring hanging here all summer ... and the weather , '' \corefboxc{Soul} nearly gasps , dramatizing it for added effect , `` Oh , man , \corefboxc{I} do n't know how \corefboxa{you} can stay cooped up in that room of \corefboxa{yours} when it 's so nice out , '' \corefboxc{he} grins .
  
  `` But ... meh .
  Different strokes .
  \corefboxc{I} ca n't judge . ''

  \corefboxc{His} comments comfort \corefboxa{them} , an for a moment \corefboxa{they} forget how this came to be .
  The cathedral in Italy , \corefboxi{Lady Medusa} 's wrath , and the black blood that infected \corefboxc{him} .
  Every moment \corefboxa{they} spent in the presence of \corefboxc{Soul Evans} builds always up to this ; fixation on the memories of \corefboxj{their} first encounters and all the pain \corefboxa{they} 've caused \corefboxc{him} , the pain \corefboxa{they} 've caused \corefboxk{\corefboxc{he} and \corefboxe{Maka}} both .
  As quickly as \corefboxc{Soul} had lifted \corefboxa{the swordsman} 's spirits , \corefboxa{they} 'd weighed \corefboxa{themselves} down once more .
  It seemed so normal , though .
  \corefboxc{Soul} could n't bring \corefboxc{himself} to feel any sense of accomplishment in the coaxing - out of \corefboxa{Crona} 's smile when the return of \corefboxa{their} self doubt was as certain as the sun in the sky .
  \corefboxc{His} own stubbornness could n't let \corefboxc{his} diminished self worth lie .
  With another encouraging smile , rows of sharpened incisors appearing oddly charismatic , \corefboxc{he} opens \corefboxc{his} mouth to speak -- but finds \corefboxc{himself} cut off before \corefboxc{he} can even squeeze a word in .
  
  `` \corefboxc{Soul} , \corefboxa{I} 'm sorry , '' \corefboxa{the meister} blurts .

  Having been pent - up for months , the apology comes forth without inhibition , rolling effortlessly off \corefboxa{their} tongue .
  
  `` Sorry ... ?
  For what ? ''
  \corefboxc{Evans} quirks a brow , chuckling .
  
  \corefboxc{He} adjusts \corefboxc{his} stance to face \corefboxa{Crona} with the whole of \corefboxc{his} body , maintaining \corefboxc{his} positive demeanor .
  
  `` F - for what ... ? ''
  
  \corefboxa{They} stammer , shaking \corefboxa{their} head .
  For all \corefboxa{their} remorse , \corefboxa{they} thought this would have been obvious .
  
  `` For everything , it 's ... the first time \corefboxf{we} dueled , \corefboxa{I} was the enemy !
  \corefboxa{I} - \corefboxa{I} almost killed \corefboxc{you} , \corefboxa{I} - \corefboxa{I} ...
  \corefboxa{I} really , really hurt \corefboxc{you} , '' \corefboxa{they} answer , still so sick with guild that even \corefboxa{their} confession of responsibility is tainted with frustration .

  \corefboxc{Soul} seems stunned for a moment before harnessing \corefboxc{his} quick wit .

  `` Hey , now , \corefboxa{you} ca n't take all the credit like that , \corefboxl{Ragnarok} did most of the damage , '' \corefboxc{he} \dots
\end{footnotesize}
